\documentclass[final]{cvpr_template/cvpr}

\usepackage{verbatim}
\usepackage{xfrac}
\usepackage{subfig}
\usepackage{mathtools}
\usepackage{times}
\usepackage{epsfig}
\usepackage{graphicx}
\usepackage{amsmath}
\usepackage{amssymb}
\usepackage{stfloats}
\usepackage{combelow}
\usepackage{flushend}
\usepackage{gensymb}
\usepackage{booktabs, multicol, multirow}

\DeclareMathOperator*{\argmin}{argmin} 

\usepackage[pagebackref=true,breaklinks=true,colorlinks,bookmarks=false]{hyperref}

\def\cvprPaperID{1188} 
\def\confYear{CVPR 2021}

\begin{document}

\title{Sparse Auxiliary Networks for Unified \\ Monocular Depth Prediction and Completion}

\author{Vitor Guizilini \quad Rare\cb{s} Ambru\cb{s} \quad Wolfram Burgard \quad Adrien Gaidon \\
Toyota Research Institute (TRI), Los Altos, CA \\
{\tt\small \{first.lastname\}@tri.global}
}

\maketitle

\begin{abstract}
Estimating scene geometry from data obtained with cost-effective sensors is key for robots and self-driving cars.
In this paper, we study the problem of predicting dense depth from a single RGB image (monodepth) with optional sparse measurements from low-cost active depth sensors.
We introduce Sparse Auxiliary Networks (SANs), a new module enabling monodepth networks to perform both the tasks of depth prediction and completion, depending on whether only RGB images or also sparse point clouds are available at inference time. First, we decouple the image and depth map encoding stages using sparse convolutions to process only the valid depth map pixels. Second, we inject this information, when available, into the skip connections of the depth prediction network, augmenting its features. Through extensive experimental analysis on one indoor (NYUv2) and two outdoor (KITTI and DDAD) benchmarks, we demonstrate that our proposed SAN architecture is able to simultaneously learn both tasks, while achieving a new state of the art in depth prediction by a significant margin.

\end{abstract}

\section{Introduction}

Dense scene geometry can be directly measured using active sensors (e.g., LiDAR, structured light) or estimated from RGB cameras (e.g., via stereo matching, structure from motion, monocular depth networks). Both approaches have complementary strengths and failure modes (e.g., rain or low light). Consequently, a robust perception system must leverage both modalities while still retaining functionality when only one is available.
In this paper, we propose a learning algorithm and model that can satisfy these desiderata with a simple sensor suite: a single monocular RGB camera combined with any low-cost active depth sensor returning only a few 3D points per scene.

\begin{figure}[t!]
\centering
\includegraphics[width=0.45\textwidth,trim=40 10 40 0, clip]{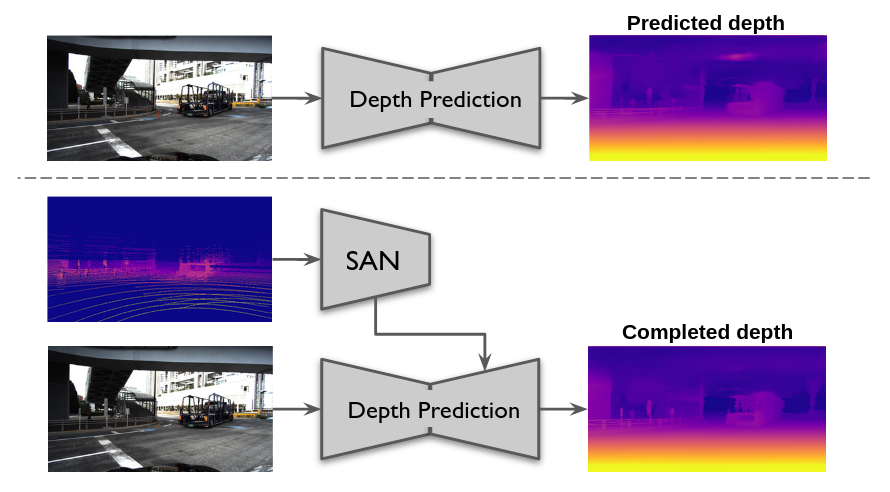}
\caption{\textbf{Our proposed joint task learning SAN architecture} produces state of the art monocular depth estimates from a single image (prediction), which can be further improved by also providing a sparse depth map (completion) \textbf{without changing the model}.}
\label{fig:teaser}
\vspace{-4mm}
\end{figure}

Monocular depth prediction is becoming a cornerstone capability for a wide range of robotic applications where RGB cameras are ubiquitous~\cite{hu2020probabilistic, watson2020footprints, yang2020d3vo}. 
Recently, self-supervised methods trained only on raw videos demonstrated that robots with a single camera can learn and predict dense depth information~\cite{bian2019unsupervised,godard2017unsupervised,monodepth2,gordon2019depth,dualnet,zhou2016learning}, especially as the quantity of data increases~\cite{packnet}. However, in practice an active range sensor is often available, and can be used to either provide further supervision at training time \cite{eigen2014depth,fu2018deep,Gan2018MonocularDE,lee2019big,Yin2019enforcing} or also during inference \cite{imran2019depth,ma2018self,qiu2019deeplidar}, in a task known as depth completion. Even though sparse, recent works \cite{packnet-semisup} have shown that even a few pixels containing valid depth information is enough to boost performance, and therefore should not be discarded.
Importantly, these two tasks, depth \emph{prediction} and \emph{completion}, are treated as separate problems with different architectures. No method to date tackles the issue of \emph{using all the information available from both modalities at both training and inference time}, including if only partially available (e.g., due to sensor blackout, occlusion, or environmental conditions). 

Our \textbf{main contribution} is a novel architecture, \textit{Sparse Auxiliary Networks} (SANs, cf.~Fig.~\ref{fig:diagram}), that enables a monocular depth \emph{prediction} network to also perform depth \emph{completion} in the presence of optional sparse 3D measurements at inference time. Note that the \emph{same architecture and weights} can \emph{dynamically} perform either task at inference time, depending on the presence or not of sparse depth measurements.
Our model relies on a sparse depth convolutional encoder to inject depth information, when available, into the skip connections of state-of-the-art encoder-decoder networks for depth prediction. 
Our \textbf{second contribution} is a thorough experimental evaluation on three challenging outdoor (KITTI~\cite{Geiger2012CVPR} and DDAD~\cite{packnet}) and indoor (NYUv2~\cite{silberman2012indoor}) datasets, demonstrating that our SAN architecture \emph{boosts monocular depth prediction performance and sets a new state of the art in this task}.

\section{Related Work}
\subsection{Depth Prediction}

Monocular depth prediction has been gaining in popularity in the robotics community, with methods generally falling into different categories depending on the data used to derive the learning signal. \textit{Self-supervised learning} methods aim to predict depth directly from monocular images, by imposing a photometric loss on temporally adjacent frames~\cite{zhou2017unsupervised} or on corresponding stereo images~\cite{godard2017unsupervised}. Owing to its simplicity and wide availability of raw data, a wide range of body of work has addressed this topic, combining it with optical flow~\cite{yin2018geonet,zhao2020towards}, uncertainty estimation~\cite{poggi2020uncertainty}, semantic segmentation~\cite{guizilini2020semantically,tosi2020distilled}, instance segmentation~\cite{bian2019unsupervised},  keypoint estimation~\cite{tang2019self} and visual odometry~\cite{yang2020d3vo,yang2018deep}. 

By contrast, \textit{supervised learning} methods apply a regression loss using ground truth depth supervision, either by minimizing mean squared error~\cite{eigen2014depth} or through ordinal regression~\cite{fu2018deep}. In addition to the standard regression loss, methods additionally use planar patches as guidance~\cite{lee2019big}, impose 3D geometric constraints~\cite{Yin2019enforcing}, use surfasce normals as regularization~\cite{qi2018geonet,wei2019enforcing}, exploit task consistency constraints between depth, normals and semantic segmentation~\cite{zhang2019pattern}, or use semantic guidance~\cite{klingner2020self,ochs2019sdnet}. A number of methods use Structure-from-Motion~\cite{klodt2018supervising} or distil stereo information~\cite{guo2018learning} to use as supervisory signal during training. Our method is conceptually similar to~\cite{guizilini2020semantically}, where the authors use pixel-adaptive convolutions to distill features from a semantic segmentation network. Instead, we propose novel \textit{Sparse Residual Blocks} which leverage Minkowski convolutions~\cite{minkowski} and are specifically designed to account for the sparse nature of our supervisory signal.

\subsection{Depth Completion}

While a high number of methods exists that focus purely on depth data, ranging from bilateral filters~\cite{tomasi1998bilateral} to recent CNN densification methods~\cite{uhrig2017sparsity}, we will focus on methods that rely on RGB images as additional information. In the case when the depth signal is sparse (e.g., LiDAR), methods typically rely on RGB-based appearance as guidance and additionally devise custom convolutions and propagate confidence to consecutive layers~\cite{propconf}, or use content-dependent and spatially-variant guiding convolutions~\cite{tang2019learning}. Alternative sources of information including confidence masks and object cues~\cite{wvangansbeke_depth_2019} as well as exploiting cross-attention between the RGB and depth encoders~\cite{lee2020deep} can also be used. 

To avoid depth mixing typically induced by the standard MSE loss a binned depth representation trained using a cross-entropy loss has been shown to work~\cite{imran2019depth}. When additional temporally-adjacent frames are available a proxy photometric loss can be derived to further constrain densification~\cite{ma2018self,zhang2019dfinenet}, while in this setting as little as 4 LiDAR beams are enough to provide a meaningful supervisory signal~\cite{packnet-semisup}. Note that our proposed method does not explicitly model any relationship between the two input modalities (RGB and depth), but rather learns these at a feature level. 
\section{Methodology}

\subsection{Monocular Depth Estimation}
\label{sec:notation}

\textbf{Prediction.}
The aim of monocular depth prediction is to learn a function $f_P:I \to D$ that takes as input image $I$ and recovers a \textit{predicted} depth $\hat{D}_P=f_P\left(I(p)\right)$ for every pixel $p\in I$ (i.e., a dense depth map).  In a supervised setting, we have access to sparse ground truth depth $D$ at training time, as acquired by an independent sensor and projected back onto the camera's image plane. Thus we treat monocular depth estimation purely as a regression problem, and learn an estimator $f_P$ parameterized by $\theta_P = \{\theta_{I}\}$ by solving:
\begin{equation} \label{eq:depth_model}
\hat{\theta}_P = \argmin_{\theta_P}\mathcal{L}_{sup}\left(f_P(I; \theta_P),D \right).
\end{equation}

\textbf{Completion.} In the monocular depth completion task, we also have access to sparse ground truth depth $\tilde{D}$ during inference (usually a subset of $D$~\cite{silberman2012indoor}, or collected by noisier/sparser sensors~\cite{geiger2013vision}). This information can be used in conjunction with $I$ to generate a \textit{completed} dense depth map $\hat{D}_C = f_C(I(p),\tilde{D}(p))$, where $f_C$ is an estimator parameterized by $\theta_{C} = \{\theta_{I}, \theta_D\}$ learned by solving:
\begin{equation} \label{eq:depth_model_completion}
\hat{\theta}_C = \argmin_{\theta_C}\mathcal{L}_{sup}\left(f_C(I,\tilde{D}; \theta_C),D \right).
\end{equation}
Note that $f_C$ contains $f_P$, in the sense that it uses the same parameters $\theta_I$ to process the input image $I$, while incorporating $\theta_D$ to process the input depth map $\tilde{D}$. This design choice is one of the core insights of this paper, as it enables feature sharing between the tasks of depth prediction and completion (see Fig.~\ref{fig:diagram}). 

\textbf{Training Loss.}
Our supervised objective is the \textit{Scale-Invariant Logarithmic} loss (SILog)~\cite{eigen2014depth}, composed by the sum of the variance and the weighted squared mean of the error in log space $\Delta d = \log d - \log \hat{d}$:
\begin{equation}
\small
\mathcal{L}(D, \hat{D}) =
\frac{1}{N} \sum_{d \in D} \Delta d ^2 - \frac{\lambda}{N^2} \left(\sum_{d \in D} \Delta d \right)^2,
\end{equation}
where $N$ is the number of valid pixels in $D$ (invalid pixels are masked out and not considered during optimization). The coefficient $\lambda$ determines the emphasis in minimizing the variance of the error. Following previous works~\cite{lee2019big}, we use $\lambda=0.85$ in all experiments. In order to train both tasks simultaneously, we add the losses generated by both output depth maps relative to the same ground truth, so that
\begin{equation}
\mathcal{L}(D,\hat{D}_P,\hat{D}_C) = \mathcal{L}(D,\hat{D}_P) + \mathcal{L}(D,\hat{D}_C).
\end{equation}

\subsection{Sparse Auxiliary Networks (SANs)}
\label{sec:srb}

Images are dense 2D representations of the information captured by a camera, which makes convolutions a natural choice in most computer vision tasks~\cite{convcv}. Depth maps, however, are very sparse, often containing less than $1\%$ valid pixels with useful information~\cite{packnet-semisup}, thus making convolutions a sub-optimal choice as: (i) significant computational power is wasted on uninformative areas; (ii) spatial dependencies will include spurious information from these uninformative areas; and (iii) shared filters will still average loss gradients from the entire input depth map.

\begin{figure*}[t!]
\vspace{-5mm}
\centering
\subfloat[\textbf{Proposed PackNet-SAN architecture.} Residual, packing, unpacking and convolutional blocks, as well as the inverse depth layer, are detailed in \cite{packnet}. The sparse residual block is detailed in Fig. \ref{fig:diagram_b}, the sparsification and densification layers are defined in Eqs. \ref{eq:sparsify} and \ref{eq:densify}, and $\mathbf{w}$ and $\mathbf{b}$ are learnable parameters defined in Eq.~\ref{eq:params}.]
{
\includegraphics[width=0.65\textwidth]{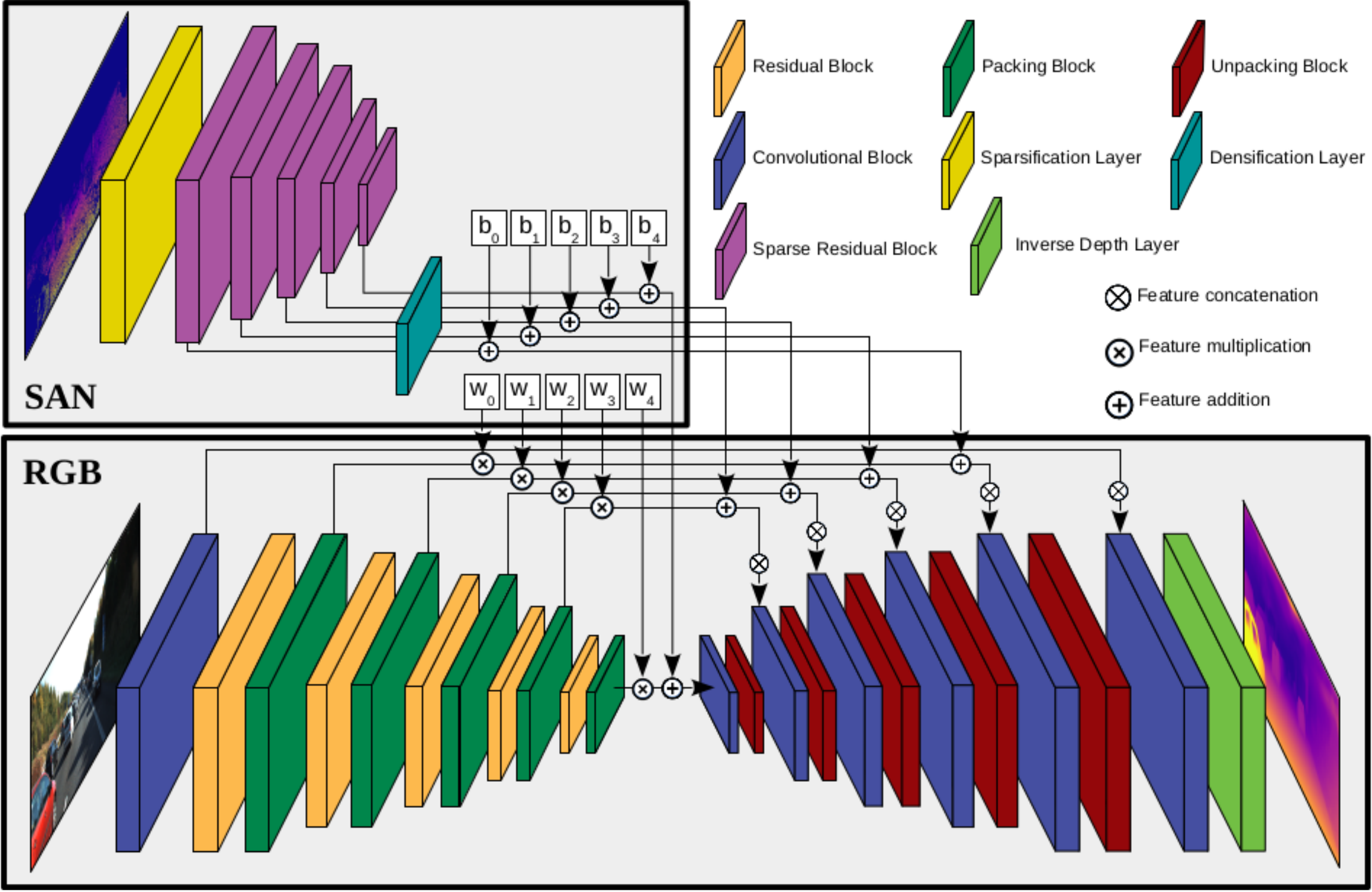}
\label{fig:diagram_a}
}
\hspace{3mm}
\subfloat[\textbf{Sparse residual block} (SRB). Each \textit{SparseConv2D} layer is a Minkowski 2D Convolution \cite{minkowski}, \textit{BN} is Batch Normalization \cite{bnorm} and \textit{ReLU} are Rectified Linear Units \cite{relu}.]{
\raisebox{7mm}{\includegraphics[width=0.29\textwidth]{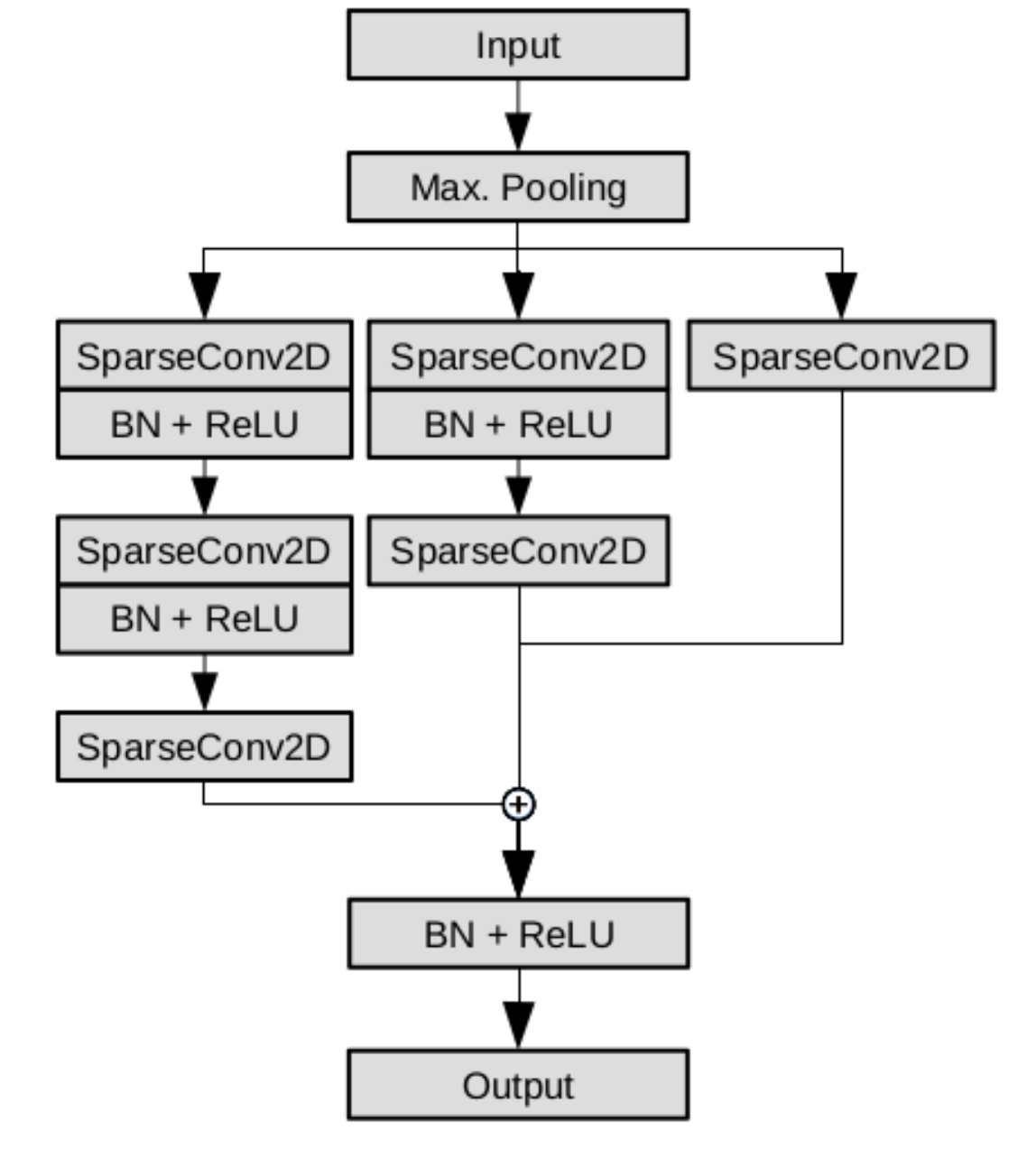}}
\label{fig:diagram_b}
}
\caption{\textbf{Our proposed SAN architecture for the joint learning of monocular depth prediction and completion}, using PackNet~\cite{packnet} as the depth prediction network (best visualized in color).}
\label{fig:diagram}
\vspace{-3mm}
\end{figure*}

To avoid these shortcomings, we propose the use of sparse convolutions to process input depth maps, while RGB images are still processed using standard convolutions. More specifically, we use Minkowski convolutions~\cite{minkowski}, a highly efficient generalized sparse convolution recently introduced to address high-dimensional problems. In this work we focus on the 2D application of Minkowski convolutions (image processing), and leave potential higher-dimension applications (i.e., multi-view~\cite{gojcic2020LearningMultiview} or spatio-temporal reasoning~\cite{minkowski}) to future work. Within this framework, a sparse tensor $S$ is written as a \textit{coordinate} matrix $C$ and a \textit{feature} matrix $F$:
\begin{equation}
C = \left[
\begin{array}{ccc}
u_1 & v_1 & s_1 \\
& \vdots & \\
u_N & v_N & s_N \\
\end{array}
\right] \quad , \quad 
F = \left[
\begin{array}{c}
\textbf{f}_1 \\
\vdots \\
\textbf{f}_N \\
\end{array}
\right],
\end{equation}
where $\{u_n, v_n\}$ are pixel coordinates, $s_n$ is the sample index in the batch, and $\textbf{f}_n \in \mathbb{R}^Q$ is the corresponding feature vector. For simplicity and without loss of generality, we assume a batch size of $1$ and disregard the batch index. An input $W \times H \times 1$ depth map $\tilde{D}$ is \textit{sparsified} by gathering its valid pixels (i.e., with positive values) as coordinates and depth values as features, such that:
\begin{align}
\small
\label{eq:sparsify}
\tilde{S} = \left\{\left\{\left(u, v\right), \tilde{D}(u, v)\right\} \:\forall\: u, v \in \tilde{D} \: | \: \tilde{D}(u, v) > 0\right\}
\end{align}
Similarly, a sparse tensor $\tilde{S}=\{\tilde{C},\tilde{F}\}$ can be \textit{densified} by scattering its pixel coordinates and feature values into a dense $W \times H \times Q$ matrix $\tilde{P}$, such that:
\begin{equation}
\label{eq:densify}
\tilde{P}(u_n, v_n) = 
\begin{cases}
\textbf{f}_n, & \text{if } \{u_n, v_n\} \in \tilde{C}.\\
\textbf{0}, & \text{otherwise}.
\end{cases}
\end{equation}

Once the input depth map is sparsified, its information is encoded through a series of novel Sparse Residual Blocks (SRB), detailed in Fig.~\ref{fig:diagram_b}. Each SRB is composed of three parallel branches that process the same input after an initial max pooling stage, each with a different number of sparse convolutional blocks. The output of these branches is added and serves as input to the next SRB at a lower spatial resolution. Note that this entire chain of operations is sparse, and thus can be performed efficiently given the usually high sparsity of projected~\cite{eigen2014depth} or sampled~\cite{silberman2012indoor} depth maps. After each block, a densification layer (Eq. \ref{eq:densify}) is used in parallel to generate a dense representation of these sparse features, that are then injected into the skip connections of the RGB module as detailed in the next section.

\subsection{Proposed Architecture}
\label{sec:architecture}

Our proposed architecture for the joint learning of monocular depth prediction and completion is depicted in Fig.~\ref{fig:diagram}. It is composed of two modules, one for the processing of dense images (RGB) and one for the processing of sparse depth maps (SAN). The dense RGB module can be any encoder-decoder depth prediction network that uses skip connections~\cite{fu2018deep,monodepth2,packnet,lee2019big}. In our work we consider two baseline state-of-the-art network architectures: \textit{PackNet}~\cite{packnet,packnet-semisup} and \textit{BTS}~\cite{lee2019big}. The sparse depth module uses our novel Sparse Residual Blocks described in Section~\ref{sec:srb} to encode sparsified depth maps used as input in conjunction with RGB images. Following the notation introduced in Section~\ref{sec:notation}, the RGB module is defined by the parameters $\theta_I$ and the depth module is defined by $\theta_D$.

If a single image $I$ is used, only the RGB module is activated and the output will be a \textit{predicted} depth map $f_P(I;\theta_I) = \hat{D}_P$. Alternatively, if sparse depth measurements $\tilde{D}$ are also provided, they serve as input to the SAN module, where they are encoded through a series of SRBs (Fig.~\ref{fig:diagram_b}) to produce sparse depth features at increasingly lower resolutions. These resolutions are designed to match those of the RGB encoder, in such a way that the sparse depth features can be injected into the dense RGB features by simply adding the two feature maps, after densification. Because the network utilizes this sparse depth information in addition to the dense RGB image, its output will be a \textit{completed} depth map $f_C(I,\tilde{D};\theta_I,\theta_D) = \hat{D}_C$.  

Empirically, we have determined that injecting this information at the skip connection level is optimal to ensure that both tasks can still be performed by the same network without degradation. In this configuration, the RGB encoder only processes image features, while the RGB decoder processes features from the RGB encoder augmented with the sparse features from the depth encoder. To further condition the skip connections and enable the switching between tasks, we also introduce learnable parameters $\textbf{w}$ and $\textbf{b}$ as part of the SAN module. Assuming $K_i$ as the feature map from the RGB encoder used as skip connection at scale resolution $i$, the augmented skip connection after introducing sparse depth information from $\tilde{P}_i$ is defined as:
\begin{equation}
\label{eq:params}
    \tilde{K}_i = w_i \times K_i + b_i + \tilde{P_i}
\end{equation}
Note that if no sparse depth information is available these parameters are not used. This enables the skip connections to be conditioned on the task being performed, and can better adapt to the introduction of additional information from the SAN module, minimizing gradient interference.
A detailed study to determine the contribution of each component of our proposed architecture can be found in \tabref{tab:depth_ablation}.

\section{Experimental Protocol}

\subsection{Implementation Details}

Our models\footnote{Code available at: ~\href{https://github.com/TRI-ML/packnet-sfm}{https://github.com/TRI-ML/packnet-sfm}} were implemented using Pytorch~\cite{paszke2017automatic} and trained across eight V100 GPUs, with batch size $b=4$ per GPU. We use the AdamW optimizer~\cite{adamw}, with $\beta_1=0.9$, $\beta_2=0.999$, starting learning rate $lr=10^{-4}$ and weight decay $wd=10^{-2}$. Our training schedule includes $30$ epochs where only the depth prediction network is trained, followed by $20$ epochs where the depth prediction encoder is frozen and only the depth completion encoder and shared decoder are trained. As training proceeds, the learning rate is decayed by a factor of $2$ after every $20$ epochs.

As the baseline depth prediction networks we considered \textit{BTS}~\cite{lee2019big} and \textit{PackNet}~\cite{packnet}, using their official Pytorch implementations. With BTS we evaluate our architecture's ability to improve upon the current state of the art in monocular depth prediction; and with PackNet we investigate whether a more complex architecture is better suited to simultaneously learn both tasks. Please note that our proposed Sparse Auxiliary Networks (SANs) can be equally applied to any other architecture, to benefit from potential improvements in speed, memory usage and performance.

\subsection{Datasets}

\textbf{KITTI.} We use the KITTI benchmark~\cite{geiger2013vision} and train on the \textit{Eigen} split, composed of 23,488 training, 888 validation and 697 testing images (from which only 652 contain accumulated ground-truth depth maps~\cite{gtkitti}). Additionally, we present results on the KITTI public leaderboard, which consists of and 500 and 1,000 frames respectively for testing depth prediction and completion methods. Following standard procedure~\cite{lee2019big}, at training time a random crop of $352 \times 704$ was used, with the addition of random horizontal flipping and color jittering.

\textbf{DDAD.} The Dense Depth for Automated Driving (DDAD)~\cite{packnet} is an urban driving dataset containing multiple synchronized cameras and depth ranges of up to 250 meters. It has a total of 12,560 training samples, from which we selected cameras $1$/$5$/$6$/$9$ for a total of 50,600 images and ground-truth depth maps. The validation set contains 3,950 samples (15,800 images) and ground-truth depth maps. Following standard procedure~\cite{packnet}, input images were downsampled to a $640 \times 384$ resolution, and for evaluation we considered distances up to 200m without any cropping. A single model was trained using all four cameras, and evaluated individually on each one.

\begin{table*}[t!]
\renewcommand{\arraystretch}{1.00}
\centering
{
\small
\setlength{\tabcolsep}{0.3em}
\begin{tabular}{l|c|ccccc|ccc}
\toprule
\multirow{2}{*}{\textbf{Method}} & \multirow{2}{*}{\textbf{Input}} &
\multicolumn{5}{|c|}{Lower is better $\downarrow$} &
\multicolumn{3}{|c}{Higher is better $\uparrow$} \\
& & 
Abs.Rel &
Sqr.Rel &
RMSE &
RMSE$_{log}$ &
SILog & 
$\delta<1.25$ &
$\delta<1.25^2$ &
$\delta<1.25^3$\vspace{0.5mm}\\
\toprule

Kuznietsov et al. \cite{kuznietsov2017semi} & RGB
& 0.113 & 0.741 & 4.621 & 0.189 & --- & 0.862 & 0.960 & 0.986
\\
Gan et al. \cite{Gan2018MonocularDE} & RGB
& 0.098 & 0.666 & 3.933 & 0.173 & --- & 0.890 & 0.964 & 0.985
\\
Guizilini et al. ~\cite{packnet-semisup} & RGB
& 0.078 & 0.378 & 3.330 & 0.121 & --- & 0.927 & --- & ---
\\
Fu et al. \cite{fu2018deep} & RGB
& 0.072 & 0.307 & \underline{2.727} & 0.120 & --- & 0.932 & 0.984 & 0.994
\\
Yin et al. \cite{Yin2019enforcing} & RGB 
& 0.072 & --- & 3.258 & 0.117 & --- & 0.938 & 0.990 & 0.998 
\\
Lee et al. ~\cite{lee2019big} & RGB
& \underline{0.059} & \underline{0.245} & 2.756 & \underline{0.096} & --- & \underline{0.956} & \underline{0.993} & \underline{0.998}

\\\cmidrule{1-10}

\multirow{2}{*}{BTS-SAN} & RGB
& 0.057 & 0.229 & 2.704 & 0.092 & 8.926 & 0.961 & 0.994 & 0.999 
\\
& RGB+D
& 0.021 & 0.038 & 1.094 & 0.037 & 3.749 & 0.996 & 0.999 & 1.000 

\\ \cmidrule{1-10}

\multirow{2}{*}{PackNet-SAN} & RGB
& \textbf{0.052} & \textbf{0.175} & \textbf{2.233} & \textbf{0.083} & \textbf{7.618} & \textbf{0.970} & \textbf{0.996} & \textbf{0.999}
\\
& RGB+D
& \textbf{0.015} & \textbf{0.028} & \textbf{0.909} & \textbf{0.032} & \textbf{3.149} & \textbf{0.997} & \textbf{0.999} & \textbf{1.000} 

\\ \cmidrule{1-10}

\textit{Improvement} & RGB
& \textit{11.9\%} & \textit{28.5\%} & \textit{18.9\%} & \textit{13.5\%} & --- & \textit{1.4\%} & \textit{0.0\%} & \textit{0.0\%}
\\

\bottomrule

\end{tabular}
}
\caption{\textbf{Depth estimation results on the KITTI dataset}, for the \textit{Eigen} test split \cite{eigen2014depth} and distances up to 80m. The \textit{Improvements} row indicate the percentual improvement between our best model (PackNet-SAN) and the current state of the art (BTS, by Lee et al. \cite{lee2019big}, underlined).} 
\label{tab:depth_kitti}
\vspace{-5mm}
\end{table*}

\textbf{NYUv2.} To evaluate our proposed methodology on other domains, we also provide results on the NYUv2 dataset~\cite{silberman2012indoor}. It consists of RGB+D data collected from 464 scenes, with 249 used for training and 215 for testing. We follow~\cite{lee2019big} and sample frames evenly from the training sequences, generating roughly \textit{36k} training RGB+D images. For depth prediction we train on images of size $640\times 480$, while for depth completion we first downsample the original frames by half and center-crop to $304\times 228$, so as to be consistent with the protocol followed by related methods~\cite{ma2018self}. Additionally, for depth completion we use input depth maps with 200 or 500 valid points respectively, randomly sampled from the original depth images, following the standard training protocol on this dataset~\cite{ma2018self}. We upsample each test prediction to the original test image resolution and evaluate on a center crop following related work~\cite{alhashim2018high,fu2018deep,lee2019big}, using the official test split of 654 frames.

\begin{table}[t!]
\renewcommand{\arraystretch}{1.00}
\centering
{
\small
\setlength{\tabcolsep}{0.3em}
\begin{tabular}{c|l|c|ccc}
\toprule

\parbox[t]{2mm}{\multirow{13}{*}{\rotatebox[origin=c]{90}{\textit{Prediction}}}} 

& \textbf{Method} &
\textbf{SILog} &
SqRel &
AbsRel &
iRMSE \\
\toprule
& SGDepth \cite{klingner2020self}
& 15.30 & 5.00\% & 13.29\% & 15.80 \\
& SDNet \cite{ochs2019sdnet}
& 14.68 & 3.90\% & 12.31\% & 15.96 \\
& VGG26-UNet \cite{guo2018learning}
& 13.41 & 2.86\% & 10.60\% & 15.06 \\
& PAP \cite{zhang2019pattern}
& 13.08 & 2.72\% & 10.27\% & 13.95 \\
& VNL \cite{wei2019enforcing}
& 12.65 & 2.46\% & 10.15\% & 13.02 \\
& SORD \cite{diaz2019soft}
& 12.39 & 2.49\% & 10.10\% & 13.48 \\
& RefinedMPL \cite{vianney2019refinedmpl}
& 11.80 & 2.31\% & 10.09\% & 13.39 \\
& DORN \cite{fu2018deep}
& 11.77 & 2.23\% & \textbf{8.78\%} & 12.98 \\
& BTS \cite{lee2019big}
& 11.67 & \textbf{2.21\%} & 9.04\% & \textbf{12.23} \\

\cmidrule{2-6}
 
& PackNet-SAN
& \textbf{11.54} & 2.35\% & 9.12\% & 12.38 \\

\midrule
\midrule

\parbox[t]{2mm}{\multirow{12}{*}{\rotatebox[origin=c]{90}{\textit{Completion}}}} 

& \textbf{Method} & \textbf{RMSE} &
iRMSE &
MAE &
iMAE \\
\toprule
& DCDC \cite{imran2019depth} 
& 1109.04 & 2.95 & 234.01 & 1.07 \\
& CSPN \cite{cheng2018depth} 
& 1019.64 & 2.93 & 279.46 & 1.15 \\
& Conf-Net \cite{hekmatian2019conf} 
& 962.28 & 3.10 & 257.54 & 1.09  \\
& Sparse-to-Dense \cite{ma2018self}
& 954.36 & 3.21 & 288.64 & 1.35 \\
& DFineNet \cite{zhang2019dfinenet}
& 943.89 & \textbf{1.39} & 304.17 & 1.39 \\
& CrossGuidance \cite{lee2020deep}
& 807.42 & 2.73 & 253.98 & 1.33 \\
& FusionNet \cite{wvangansbeke_depth_2019}
& 772.87 & 2.19 & \textbf{215.02} & \textbf{0.93} \\
& GuideNet \cite{tang2019learning}
& \textbf{736.24} & 2.25 & 218.83 & 0.99 \\
\cmidrule{2-6}
& PackNet-SAN
& 914.35 & 2.78 & 298.04 & 1.36 \\
\bottomrule

\end{tabular}
}
\caption{\textbf{Depth estimation results on the official KITTI testset benchmark} relative to other published methods, for both prediction and completion tasks (bold metrics are used for leaderboard scoring). Note that the same model was used in both submissions, only modifying the input information (RGB for prediction and RGB+D for completion).}
\label{tab:depth_benchmark}
\vspace{-5mm}
\end{table}

\begin{figure}[t!]
\vspace{-3mm}
\centering
\subfloat{
\includegraphics[width=0.22\textwidth]{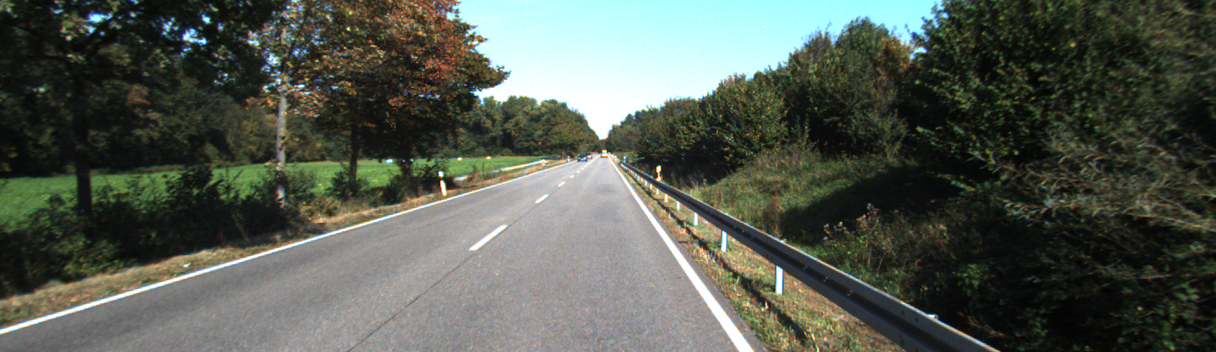}
\includegraphics[width=0.22\textwidth]{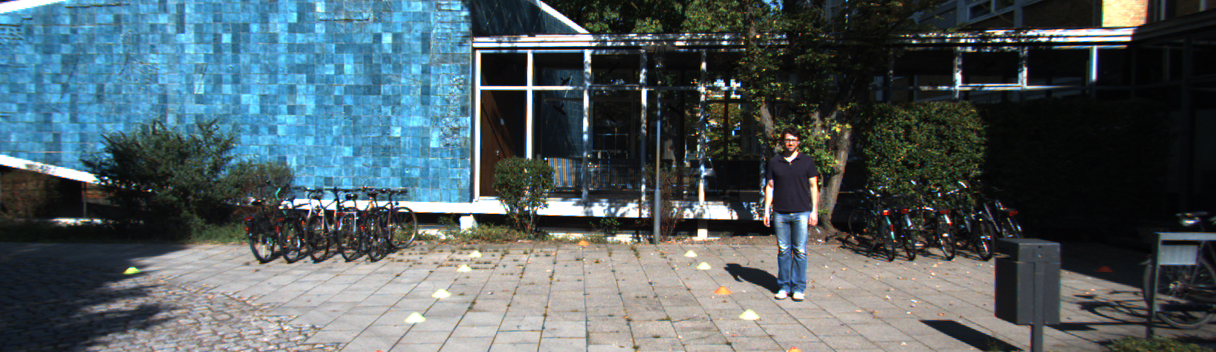}}
\\ \vspace{-4mm}
\subfloat{
\includegraphics[width=0.22\textwidth]{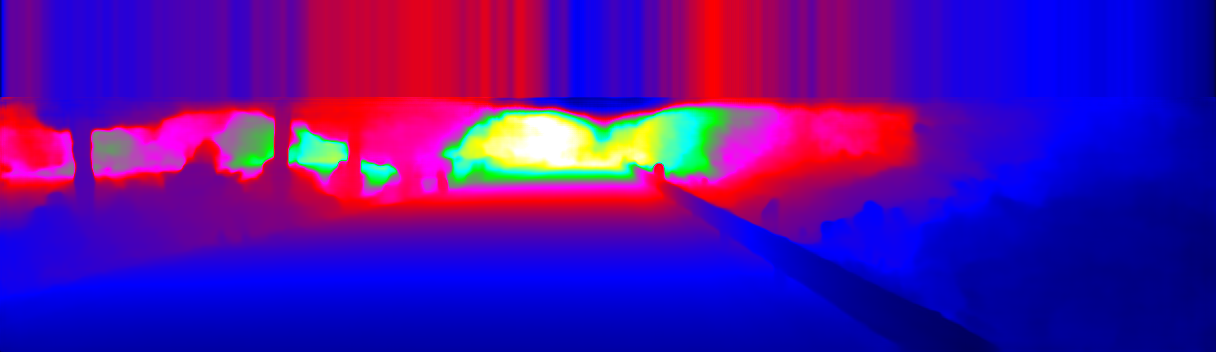}
\includegraphics[width=0.22\textwidth]{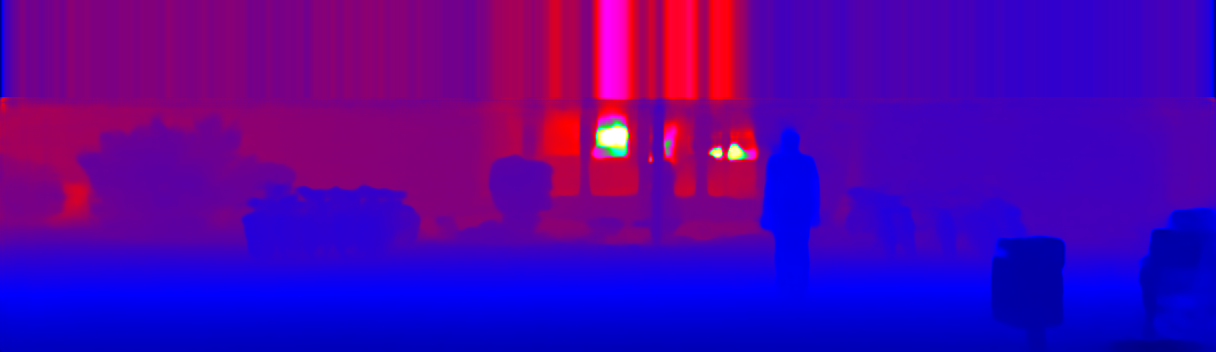}}
\\ \vspace{-4mm}
\setcounter{subfigure}{0}
\subfloat[Prediction]{
\includegraphics[width=0.22\textwidth]{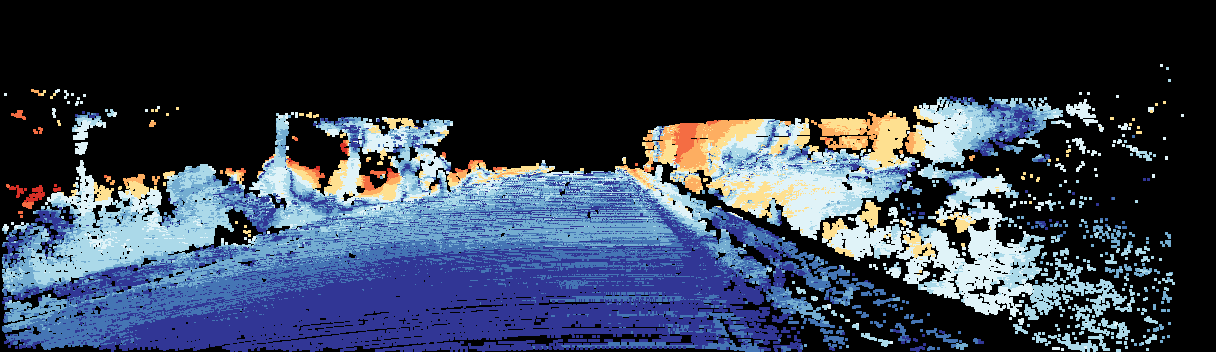}
\includegraphics[width=0.22\textwidth]{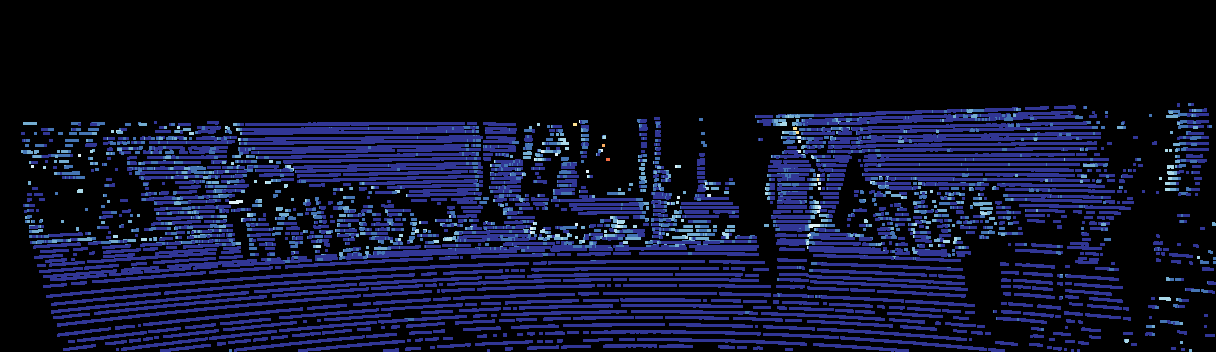}}
\\
\vspace{-3mm}
\subfloat{
\includegraphics[width=0.22\textwidth]{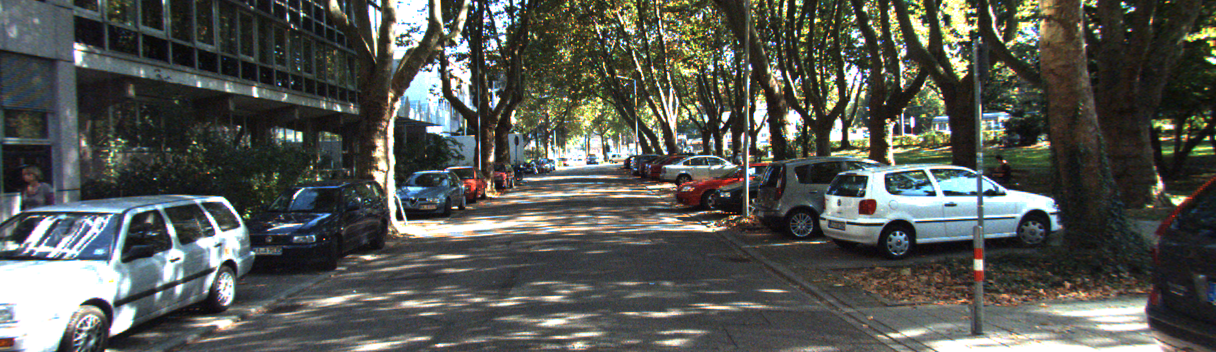}
\includegraphics[width=0.22\textwidth]{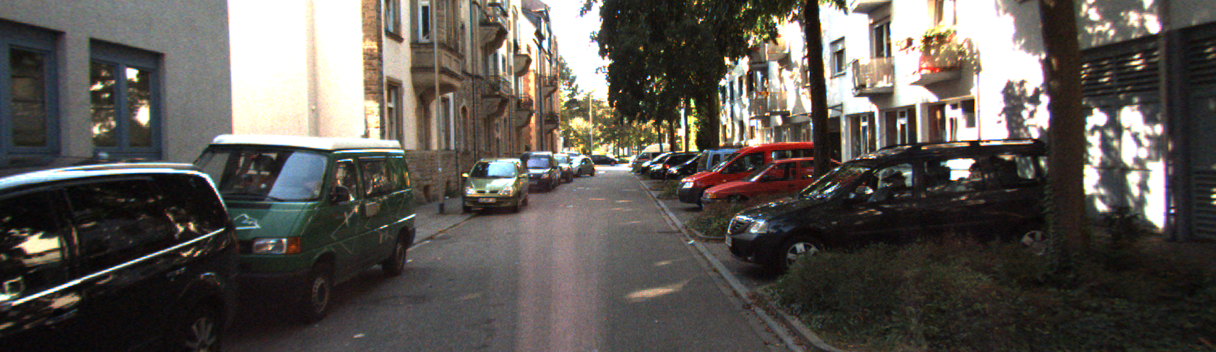}}
\\ \vspace{-4mm}
\subfloat{
\includegraphics[width=0.22\textwidth]{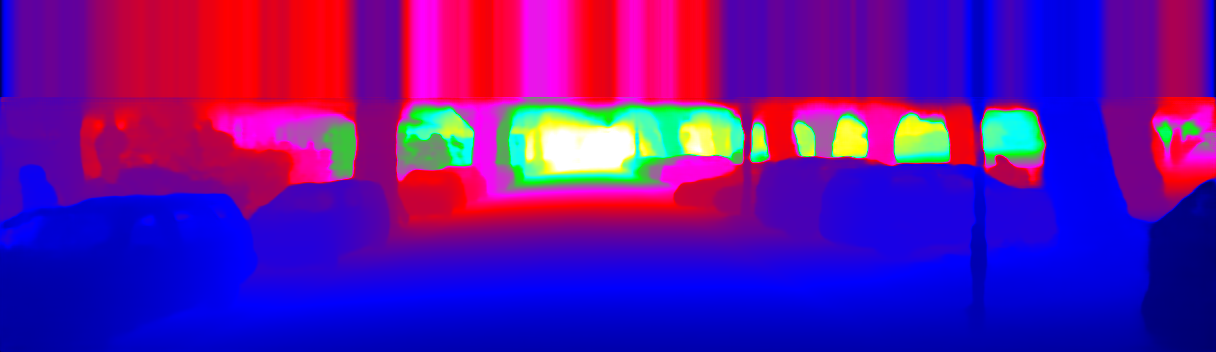}
\includegraphics[width=0.22\textwidth]{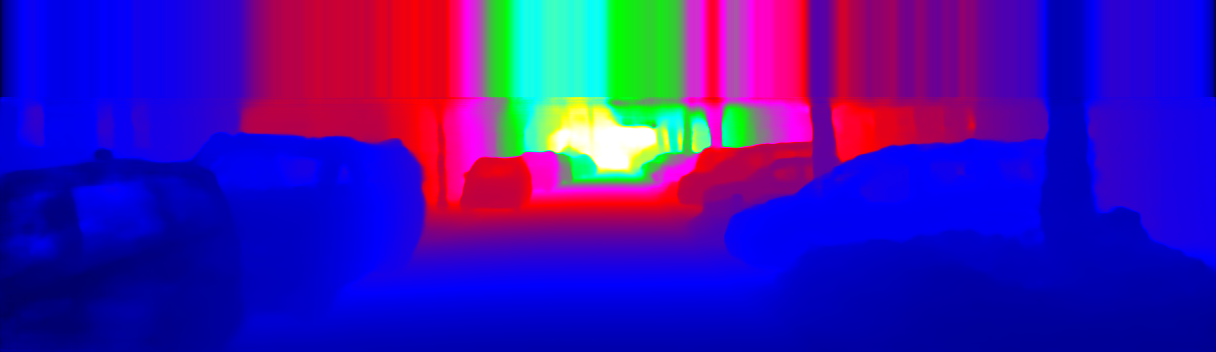}}
\\ \vspace{-4mm}
\setcounter{subfigure}{1}
\subfloat[Completion]{
\includegraphics[width=0.22\textwidth]{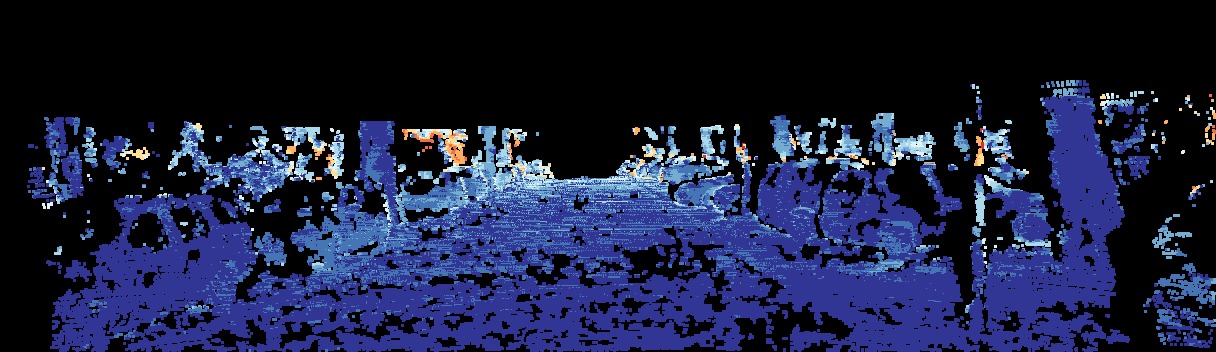}
\includegraphics[width=0.22\textwidth]{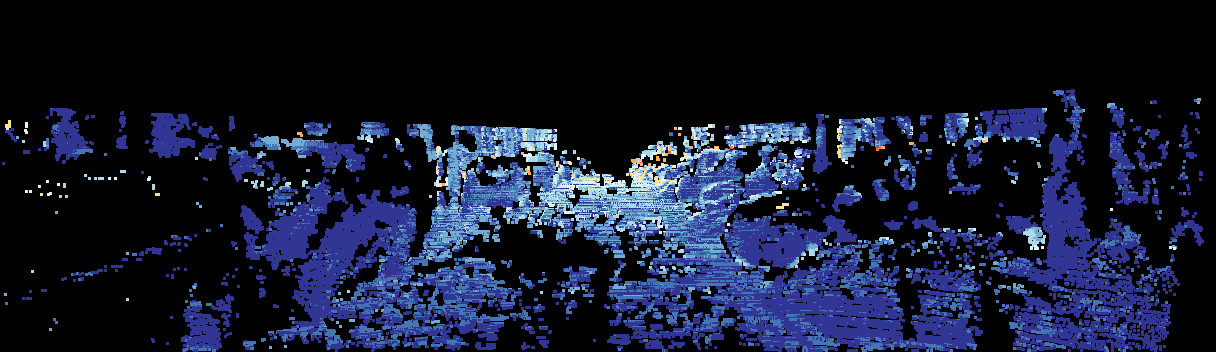}}
\caption{\textbf{Qualitative depth prediction and completion results on the KITTI benchmark}, using PackNet-SAN.}
\label{fig:depth_kitti}
\vspace{-5mm}
\end{figure}

\textbf{KITTI3D.} To further analyze the accuracy of the depth maps predicted by our proposed SAN architecture, we also evaluated their performance in the downstream task of monocular 3D object detection as \textit{pseudo-LiDAR} pointclouds. Specifically, we use the KITTI3D dataset~\cite{Geiger2012CVPR}, composed of 3,712 training and 3,712 validation images.

\textbf{Pretraining.} Following related work~\cite{lee2019big,ma2018self,wvangansbeke_depth_2019}, we found pre-training to improve network performance. For our KITTI experiments we pre-train on a larger split of DDAD, while for the NYU experiments we pre-train on the Scannet dataset~\cite{dai2017scannet} by sampling approximately $250k$ RGB+D frames without any additional cropping or filtering. We ablate the effect of pretraining in \tabref{tab:depth_ablation}.

\section{Experimental Results}
\subsection{Depth Prediction and Completion}

\textbf{KITTI}. In \tabref{tab:depth_kitti} we present quantitative results for the tasks of depth prediction and completion, considering the \textit{Eigen} test split. We note that \textit{BTS-SAN}, i.e. the BTS architecture~\cite{lee2019big} with our proposed SAN module improves over the baseline numbers for the task of depth prediction, while at the same time enabling depth completion if sparse depth maps are also provided as additional input. These results are further improved by using PackNet~\cite{packnet} as the underlying depth prediction network, establishing a new state of the art for this task by a significant margin. We also evaluated our proposed \textit{PackNet-SAN} architecture on the official KITTI test set benchmark, submitting results from the same model to both depth prediction and completion leaderboards (see \tabref{tab:depth_benchmark}). Despite operating in this challenging setting, at the time of publication our method ranked first amongst all published methods for the task of depth prediction for the \textit{SILog} metric (used to determine ranking), while at the same time showing good depth completion performance. We show qualitative results obtained from the KITTI leaderboard in Fig.~\ref{fig:depth_kitti}.

\begin{table}[t!]
\vspace{-3mm}
\renewcommand{\arraystretch}{0.9}
\centering
{
\small
\setlength{\tabcolsep}{0.3em}
\begin{tabular}{l|c|cccc}
\toprule
\textbf{Method} & \textbf{Input} &
Abs.Rel$\downarrow$ &
RMSE$\downarrow$ &
SILog$\downarrow$ & 
$\delta<1.25$$\uparrow$ \\
\toprule

\multirow{2}{*}{SRB x1} 
& RGB
& 0.057 & 2.483 & 8.064 & 0.966  
\\
& RGB+D
& 0.019 & 0.994 & 3.343 & \textbf{0.997}  
\\
\midrule
\multirow{2}{*}{SRB x2} 
& RGB
& 0.055 & 2.328 & 7.862 & 0.966  
\\
& RGB+D
& 0.017 & 0.949 & 3.287 & \textbf{0.997}
\\
\midrule
Unfreeze
& RGB
& 0.055 & 2.306 & 7.978 & 0.967  
\\
Pred. Encoder & RGB+D
& 0.021 & 0.965 & 3.333 & 0.996  
\\
\midrule
Freeze
& RGB
& 0.054 & 2.318 & 7.901 & 0.968  
\\
Pred. Decoder & RGB+D
& 0.024 & 1.070 & 3.805 & 0.995
\\
\midrule
W/o $W_i$ and
& RGB
& 0.056 & 2.374 & 8.324 & 0.962  
\\
$B_i$ parameters & RGB+D
& 0.019 & 0.958 & 3.395 & 0.995 
\\
\midrule
Train from
& RGB
& 0.062 & 2.888 & 9.579 & 0.955  
\\
Scratch & RGB+D
& 0.019 & 1.049 & 3.631 & 0.996  
\\
\midrule
Prediction
& RGB
& 0.054 & 2.476 & 8.081 & 0.966  
\\
\midrule
Completion
& RGB+D
& \textbf{0.015} & \textbf{0.878} & 3.238 & \textbf{0.997}  
\\
\midrule
\multirow{2}{*}{\textbf{PackNet-SAN}} 
& RGB
& \textbf{0.052} & \textbf{2.233} & \textbf{7.618} & \textbf{0.970} 
\\
& RGB+D
& \textbf{0.015} & 0.909 & \textbf{3.149} & \textbf{0.997}  
\\
\bottomrule

\end{tabular}
}
\caption{\textbf{Ablation analysis on the KITTI dataset}, considering the \textit{Eigen} test split \cite{eigen2014depth} and \textit{PackNet} \cite{packnet} as the depth prediction network. \textit{SRB xX} uses Sparse Residual Blocks with fewer branches; \textit{Unfreeze Pred. Encoder} also updates the prediction encoder during the second stage of training; \textit{Freeze Pred. Decoder} also freezes the prediction decoder during the second stage of training; \textit{w/o $W_i$ and $B_i$} removes the shared parameters for each skip connection; \textit{Train from scratch} does not use a pre-trained model; and \textit{Prediction} and \textit{Completion} are trained only for that particular task.}
\label{tab:depth_ablation}
\vspace{-2mm}
\end{table}

\begin{table}[b!]
\vspace{-3mm}
\renewcommand{\arraystretch}{1.00}
\centering
{
\small
\setlength{\tabcolsep}{0.3em}
\begin{tabular}{l|ccc}
\toprule
\textbf{Method} &
\textbf{AP3D@easy} &
\textbf{AP3D@medium} &
\textbf{AP3D@hard} \\
\toprule

DORN \cite{fu2018deep}
& 34.8/35.1 & 22.0/22.0 & 19.5/19.6 
\\
PackNet-SAN 
& \textbf{35.5/35.7} & \textbf{22.6/22.8} & \textbf{19.9/20.1} 
\\

\bottomrule

\end{tabular}
}
\caption{\textbf{3D object detection results on the validation set of KITI3D for the Car category}, using PatchNet \cite{Ma_2020_ECCV} and different monocular pointclouds (no input sparse depth), for the \textit{validation} split. The same detection architecture and learning hyperparameters were used in both cases.}
\label{tab:depth_detection}
\end{table}

\begin{figure*}[t!]
\small
\vspace{-5mm}
\centering
\raisebox{9mm}{\rotatebox[origin=c]{90}{\textit{RGB}}}
\subfloat{
\includegraphics[width=0.21\textwidth,height=2.0cm]{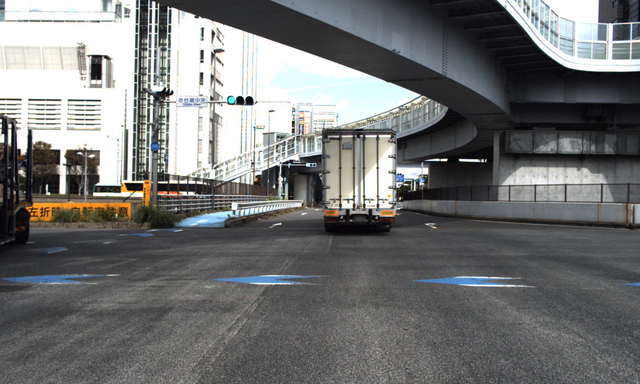}}
\subfloat{
\includegraphics[width=0.21\textwidth,height=2.0cm]{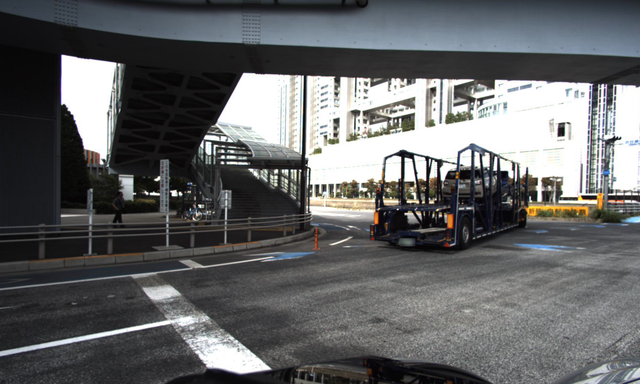}}
\subfloat{
\includegraphics[width=0.21\textwidth,height=2.0cm]{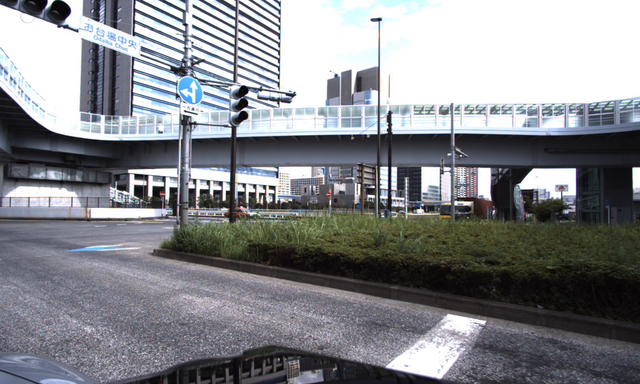}}
\subfloat{
\includegraphics[width=0.21\textwidth,height=2.0cm]{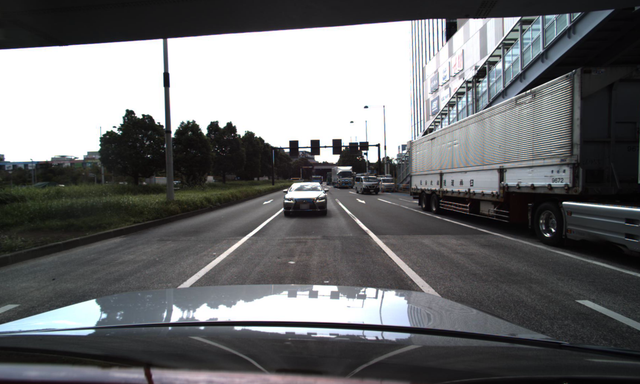}}
\\ \vspace{-4mm}
\raisebox{10mm}{\rotatebox[origin=c]{90}{\textit{Prediction}}}
\subfloat{
\includegraphics[width=0.21\textwidth,height=2.0cm]{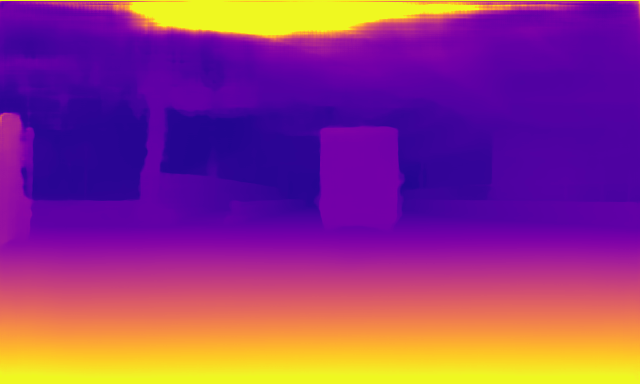}}
\subfloat{
\includegraphics[width=0.21\textwidth,height=2.0cm]{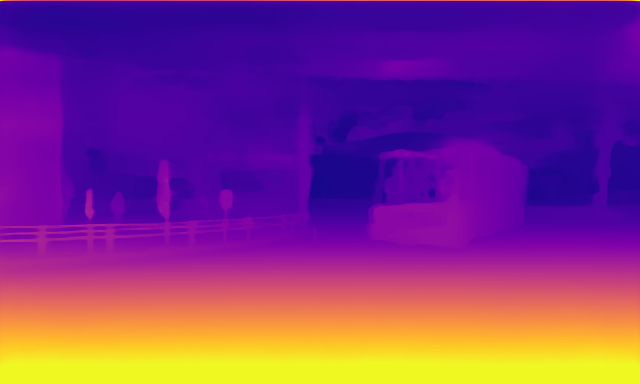}}
\subfloat{
\includegraphics[width=0.21\textwidth,height=2.0cm]{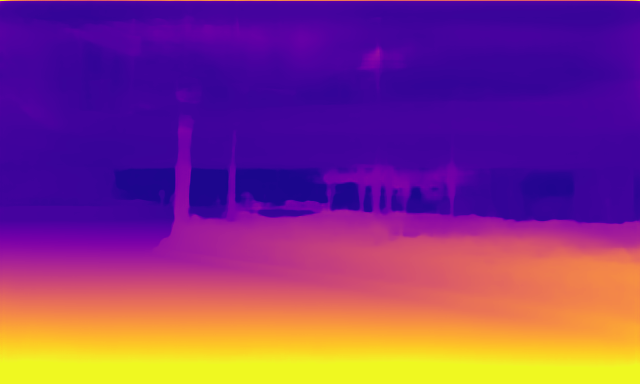}}
\subfloat{
\includegraphics[width=0.21\textwidth,height=2.0cm]{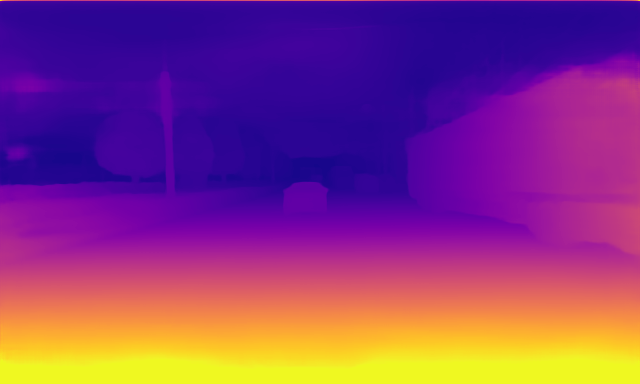}}
\\ \vspace{-4mm}
\raisebox{10mm}{\rotatebox[origin=c]{90}{\textit{Completion}}}
\subfloat{
\includegraphics[width=0.21\textwidth,height=2.0cm]{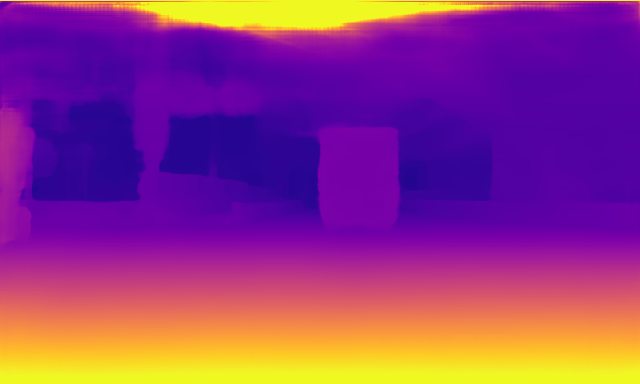}}
\subfloat{
\includegraphics[width=0.21\textwidth,height=2.0cm]{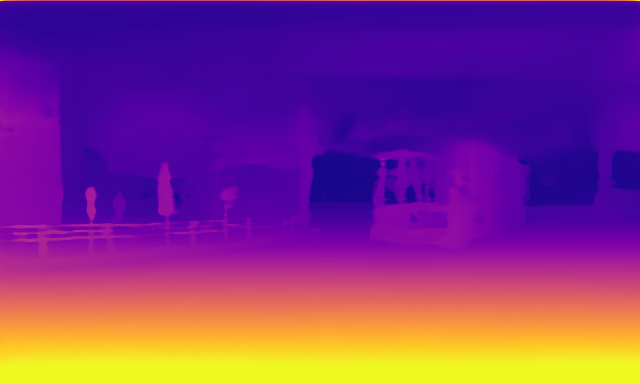}}
\subfloat{
\includegraphics[width=0.21\textwidth,height=2.0cm]{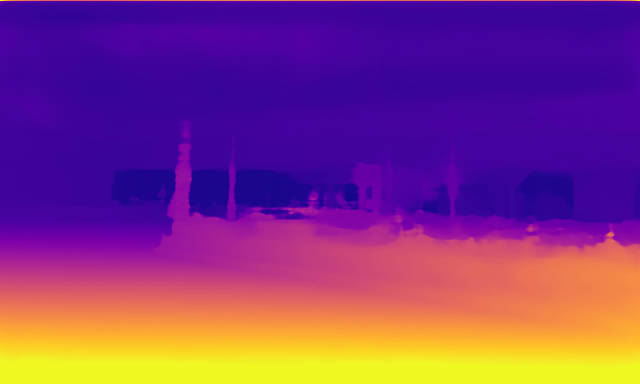}}
\subfloat{
\includegraphics[width=0.21\textwidth,height=2.0cm]{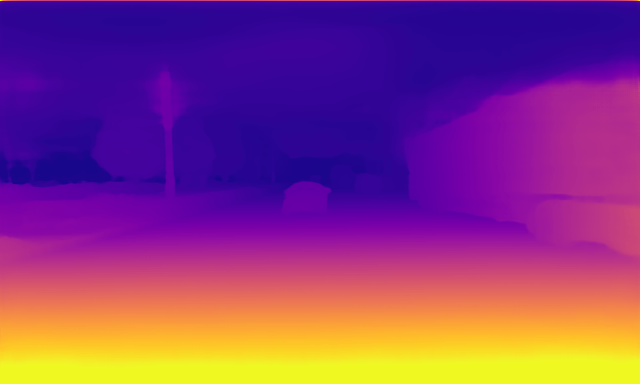}}
\setcounter{subfigure}{0}
\caption{\textbf{Qualitative depth results on DDAD}, using PackNet-SAN. A single sample is shown, each column corresponding to an individual camera. The same model was trained on all four cameras both for the task of depth prediction (middle row) and completion (bottom row), as shown in  \tabref{tab:depth_ddad}.}
\label{fig:qualitative_ddad}
\vspace{-4mm}
\end{figure*}

\begin{figure}[t!]
\vspace{-3mm}
\centering
\includegraphics[width=0.49\textwidth]{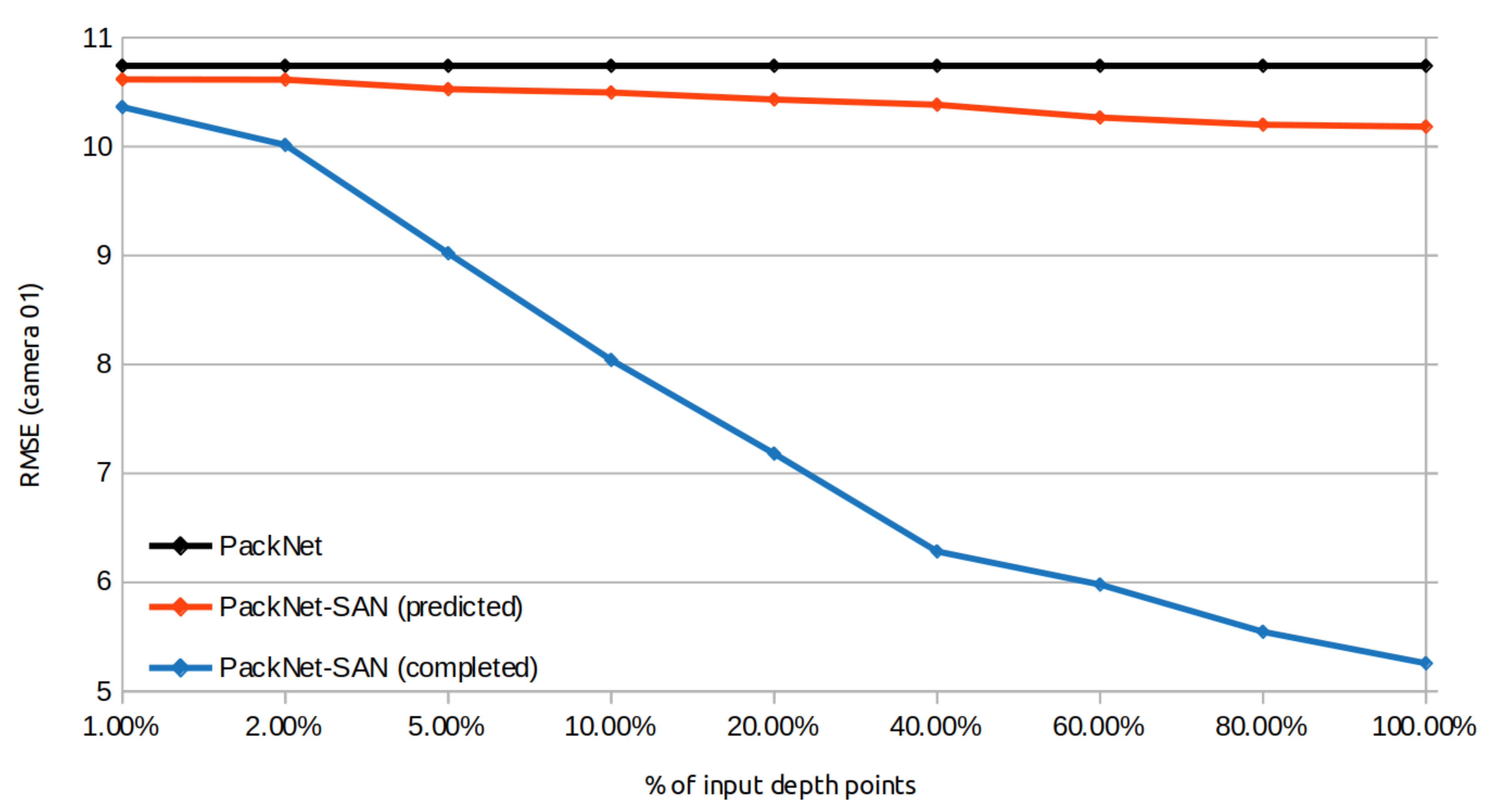}
\caption{\textbf{Sparsity analysis on the DDAD dataset}, using PackNet-SAN. We show depth prediction (red) and completion (blue) results with different levels of sparsity, compared to the baseline prediction-only network (black).}
\label{fig:ddad_sparsity}
\end{figure}

\textbf{DDAD.} In \tabref{tab:depth_ddad} we show results on the DDAD dataset obtained using our baseline depth prediction network (PackNet) and its extension using our proposed SAN architecture, to enable the joint-task learning of depth prediction and completion. From these results we note that the introduction of joint-task learning boosted depth prediction results by a significant margin, similarly to
what was observed in the KITTI experiments (for qualitative results, see Fig.~\ref{fig:qualitative_ddad}). Additionally, if sparse depth maps are available as input they can also be used to generate depth completion results, further improving performance. We note that the RGB+D experiments on DDAD were performed with a sparsity level of 20\% for the input depth maps - we provide a detailed analysis of how the sparsity level affects performance in the ablative section (see Fig.~\ref{fig:ddad_sparsity}).

\begin{table*}[t!]
\renewcommand{\arraystretch}{0.9}
\centering
{
\small
\setlength{\tabcolsep}{0.3em}
\begin{tabular}{c|c|c|ccccc|ccc}
\toprule
& \multirow{2}{*}{\textbf{Input}} & \multirow{2}{*}{\textbf{Camera}} &
\multicolumn{5}{|c|}{Lower is better $\downarrow$} &
\multicolumn{3}{|c}{Higher is better $\uparrow$} \\
& & & 
Abs.Rel &
Sqr.Rel &
RMSE &
RMSE$_{log}$ &
SILog & 
$\delta<1.25$ &
$\delta<1.25^2$ &
$\delta<1.25^3$\vspace{0.5mm}\\
\toprule

\parbox[t]{2mm}{\multirow{4}{*}{\rotatebox[origin=c]{90}{\textit{PackNet}}}} 
& \multirow{4}{*}{RGB} 
& 01 & 0.088 & 1.760 & 11.331 & 0.195 & 18.499 & 0.899 & 0.960 & 0.981 
\\
& & 05 & 0.130 & 2.025 & 10.472 & 0.268 & 25.273 & 0.832 & 0.927 & 0.960
\\
& & 06 & 0.151 & 2.485 & 10.680 & 0.307 & 28.007 & 0.791 & 0.904 & 0.944
\\
& & 09 & 0.132 & 2.362 & 12.497 & 0.261 & 24.551 & 0.821 & 0.925 & 0.962
\\
\midrule
\midrule
\parbox[t]{2mm}{\multirow{9}{*}{\rotatebox[origin=c]{90}{\textit{PackNet-SAN}}}} 
& \multirow{5}{*}{RGB} 
& 01 & 0.083 & 1.575 & 10.693 & 0.185 & 17.767 & 0.911 & 0.967 & 0.987 
\\
& & 05 & 0.127 & 1.863 & 10.210 & 0.263 & 24.966 & 0.841 & 0.931 & 0.973
\\
& & 06 & 0.145 & 2.307 & 10.493 & 0.298 & 27.491 & 0.804 & 0.911 & 0.968
\\
& & 09 & 0.119 & 1.979 & 12.010 & 0.256 & 24.295 & 0.844 & 0.936 & 0.978
\\
\cmidrule{3-11}
& & {\textit{Avg. Improv.}} & \textit{5.45\%} & \textit{10.47\%} & \textit{3.44\%} & \textit{2.96\%} & \textit{2.01\%} & \textit{1.71\%} & \textit{0.78\%} & \textit{1.54\%}
\\
\cmidrule{2-11}
\cmidrule{2-11}
& \multirow{4}{*}{RGB+D} 
& 01 & 0.052 & 0.933 & 8.683 & 0.153 & 14.920 & 0.955 & 0.978 & 0.987 
\\
& & 05 & 0.072 & 1.097 & 7.950 & 0.207 & 20.375 & 0.928 & 0.958 & 0.973
\\
& & 06 & 0.081 & 1.255 & 7.994 & 0.232 & 22.675 & 0.922 & 0.955 & 0.969
\\
& & 09 & 0.067 & 1.131 & 9.052 & 0.189 & 18.481 & 0.934 & 0.966 & 0.979
\\
\bottomrule

\end{tabular}
}
\caption{\textbf{Depth estimation results on DDAD} using PackNet-SAN, with the same model trained on four cameras considering distances up to 200m. For the RGB+D experiments, a sparsity level of 20\% was used for input depth maps (see Fig. \ref{fig:ddad_sparsity}).} 
\label{tab:depth_ddad}
\end{table*}

\begin{figure}[t!]
\vspace{-3mm}
\centering
\subfloat{
\includegraphics[width=0.15\textwidth,height=2.0cm]{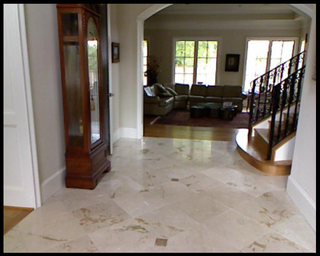}}
\subfloat{
\includegraphics[width=0.15\textwidth,height=2.0cm]{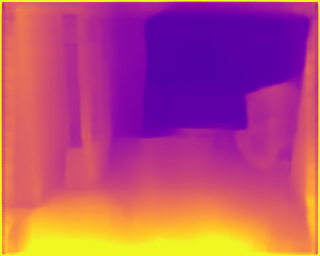}}
\subfloat{
\includegraphics[width=0.15\textwidth,height=2.0cm]{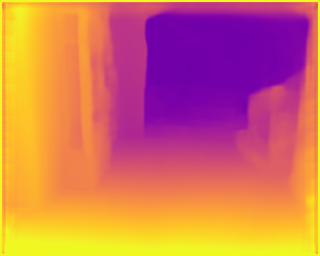}}
\\ \vspace{-3mm}
\subfloat{
\includegraphics[width=0.15\textwidth,height=2.0cm]{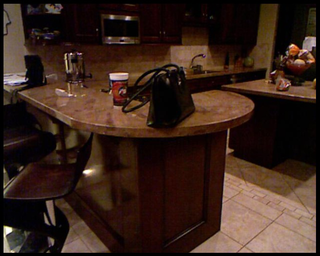}}
\subfloat{
\includegraphics[width=0.15\textwidth,height=2.0cm]{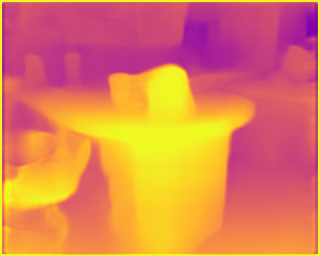}}
\subfloat{
\includegraphics[width=0.15\textwidth,height=2.0cm]{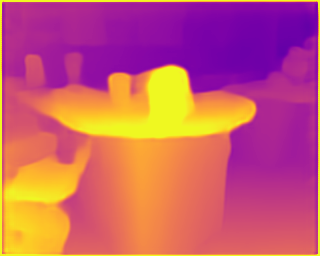}}
\\ \vspace{-3mm}
\subfloat{
\includegraphics[width=0.15\textwidth,height=2.0cm]{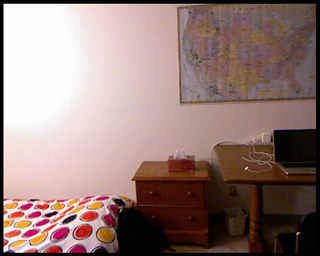}}
\subfloat{
\includegraphics[width=0.15\textwidth,height=2.0cm]{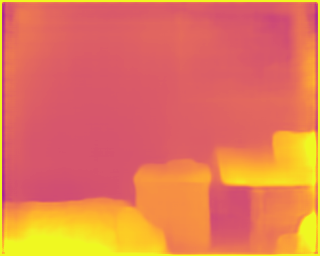}}
\subfloat{
\includegraphics[width=0.15\textwidth,height=2.0cm]{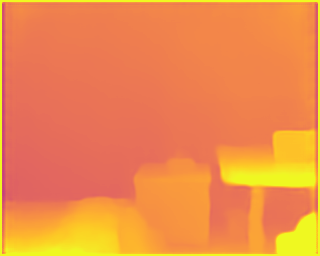}}
\\ \vspace{-3mm}
\setcounter{subfigure}{0}
\subfloat[Input]{
\includegraphics[width=0.15\textwidth,height=2.0cm]{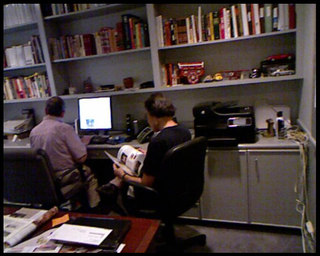}}
\subfloat[Predicted]{
\includegraphics[width=0.15\textwidth,height=2.0cm]{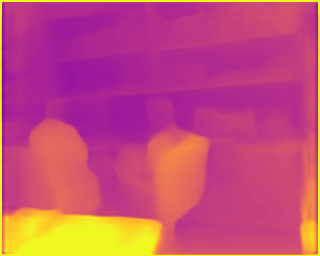}}
\subfloat[Completed]{
\includegraphics[width=0.15\textwidth,height=2.0cm]{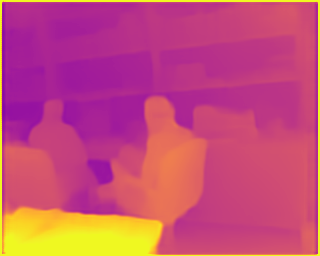}}
\caption{\textbf{Qualitative depth results on NYUv2}, using PackNet-SAN. Our joint-task learning methodology enables the generation of state-of-the-art predicted depth maps, that can be further improved by using sparse depth maps as additional input without changing the architecture.}
\label{fig:depth_nyu}
\end{figure}

\textbf{NYUv2.} Our NYUv2 results are summarized in \tabref{tab:depth_nyuv2}, observing the same trend as on the other datasets. The proposed architecture \textit{PackNet-SAN} improves over the baseline PackNet~\cite{packnet}, achieving new state-of-the-art performance for depth prediction on this dataset. When using RGB+D data at inference time, our method is competitive with state-of-the-art methods, achieving similar numbers on most metrics. We show qualitative results on NYUv2 in Fig.~\ref{fig:depth_nyu}.

\begin{table}[t!]
\renewcommand{\arraystretch}{1.00}
\centering
{
\small
\setlength{\tabcolsep}{0.15em}
\begin{tabular}{l|cc|ccc}
\toprule
\multicolumn{6}{c}{\textbf{Depth Prediction}}  \\
\midrule

\textbf{Method} &
AbsRel &
RMSE &
$\delta<1.25$ &
$\delta<1.25^2$ &
$\delta<1.25^3$ \\
\midrule
Qi et al.~\cite{qi2018geonet} & 0.128 & 0.569 & 0.834 & 0.960 & 0.990 \\
Alhashim et al.~\cite{alhashim2018high} &  0.123 & 0.465 & 0.846 & 0.974 & 0.994 \\
Fu et al.~\cite{fu2018deep} & 0.115 & 0.509 & 0.828 & 0.965 & 0.992 \\
Yin et al.~\cite{Yin2019enforcing} & 0.108 & 0.416 & 0.875 & 0.976 & 0.994 \\
Lee et al.~\cite{lee2019big} & 0.110 & \textbf{0.392} & 0.885 & 0.978 & 0.994 \\

PackNet \cite{packnet} & 0.110 & 0.397 & 0.886 & \textbf{0.979} & \textbf{0.995} \\
\midrule 
\textbf{PackNet-SAN} & \textbf{0.106} & 0.393 & \textbf{0.892} & \textbf{0.979} & \textbf{0.995} \\
\midrule
\midrule
\multicolumn{6}{c}{\textbf{Depth Completion - 200 samples}}  \\
\midrule

Ma et al.~\cite{ma2018self}$\dagger$ & 0.044 & 0.230 & 0.971 & 0.994 & 0.998 \\
NConv-CNN~\cite{propconf}$\dagger$ & 0.027 & 0.173 & 0.982 & 0.996 & 0.999 \\
Tang et al.~\cite{tang2019learning} & \textbf{0.024} & \textbf{0.142} & 0.988 & \textbf{0.998} & \textbf{1.000} \\
\midrule
\textbf{PackNet-SAN} & 0.027 & 0.155 & \textbf{0.989} & \textbf{0.998} & 0.999 \\
\midrule
\midrule
\multicolumn{6}{c}{\textbf{Depth Completion - 500 samples}}  \\
\midrule

Ma et al.~\cite{ma2018self} & 0.043 & 0.204 & 0.978 & 0.996 & 0.999 \\
DeepLidar~\cite{qiu2019deeplidar} & 0.022 & 0.115 &  0.993 & 0.999 & 1.000 \\
EncDec-Net[EF]~\cite{propconf} & 0.017 & 0.123 & 0.991 & 0.998 & 1.000 \\
CSPN~\cite{cheng2018depth} & 0.016 & 0.117 & 0.992 & 0.999 & 1.000 \\
Tang et al.~\cite{tang2019learning} & \textbf{0.015} & \textbf{0.101} & \textbf{0.995} & \textbf{0.999} & \textbf{1.000} \\
\midrule
\textbf{PackNet-SAN} & 0.019 & 0.120 & 0.994 & \textbf{0.999} & \textbf{1.000} \\
\bottomrule

\end{tabular}
}
\caption{\textbf{Depth estimation results on the test split of the NYUv2 dataset.} relative to other published methods, for both depth prediction and completion tasks. Note that the same model was used in both submissions, the only modification being the input information (RGB for prediction and RGB+D for completion). $\dagger$ - results from~\cite{tang2019learning}.}
\label{tab:depth_nyuv2}
\end{table}

\subsection{Monocular 3D Object Detection} 

To further analyze the accuracy of the depth maps predicted by our proposed SAN architecture, we evaluated their performance in the downstream task of monocular 3D object detection, using the recently proposed PatchNet architecture~\cite{Ma_2020_ECCV}. The depth maps predicted by \textit{PackNet-SAN} were projected into 3D as \textit{pseudo-LiDAR} pointclouds using ground-truth camera intrinsics, and then used as input to PatchNet without any other modifications. In \tabref{tab:depth_detection} we present results on the KITTI3D dataset and show that by operating on our pointclouds we increase object detection performance in all difficulty thresholds relative to the previous state of the art when using the DORN~\cite{fu2018deep} depth estimator. Note that, for a fair comparison, the two depth networks (DORN and PackNet-SAN) were trained using the same \textit{Eigen} split of KITTI. 

\subsection{Ablative Analysis}

In \tabref{tab:depth_ablation} we perform a comprehensive ablation study showing the effects of each design choice of our proposed architecture, and how they contribute to these improvements. In particular we show that the joint learning of both tasks actually improves depth prediction performance relative to single task learning, without degrading depth completion performance. We also demonstrate that increasing the complexity of the sparse encoder (i.e. introducing more sparse residual blocks) benefits both tasks, since it facilitates the decoupling of RGB and depth features without overloading the shared decoder. We also experimented with different schedules for parameter freezing, and determined that freezing the dense encoder after the initial depth prediction learning stage leads to optimal results.  

Additionally, in Fig.~\ref{fig:ddad_sparsity} we analyze the impact of sparsity in the input depth maps on the \textit{DDAD} dataset. Specifically, we sparsify the input depth maps by randomly sampling a percentage of valid input depth pixels (depth maps used for supervision and evaluation were not modified). As expected, performance increases with the percentage of available input depth points, showing that our proposed SAN architecture is able to leverage different levels of sparsity to consistently improve results. Interestingly, we observe a similar trend for depth prediction results as well, as further evidence that joint-task learning of depth prediction and completion is able to further improve results even when only RGB images are utilized at test time.

\section{Conclusion}
This paper describes a novel methodology for monocular depth estimation that combines the tasks of depth prediction and completion into a single architecture. We propose a mid-level fusion approach for the joint learning of both tasks, using a standard depth prediction network with the addition of a sparse encoder to process input depth maps. The sparse depth features are added to the skip connections of the image encoder at each layer, before they are fed into a shared dense decoder. The resulting architecture can be used to perform both tasks without any further training, simply by modifying the input information between RGB and RGB+D or controlling the sparsity level of the input depth maps. Through an extensive analysis on different benchmarks, we demonstrate that our proposed unified SAN architecture achieves a new state of the art in monocular depth prediction by a significant margin. As future work, we will explore multi-frame extensions (e.g., stereo pairs or temporal context), as well as developing ways to further improve depth completion performance in the SAN setting.

{\small
\bibliographystyle{cvpr_template/ieee_fullname}
\bibliography{references}
}

\end{document}